%% file: main.tex
\documentclass[letterpaper, 10 pt, conference]{ieeeconf}  

\IEEEoverridecommandlockouts                              

\usepackage{times}
\usepackage{multicol}
\usepackage[bookmarks=true]{hyperref}
\usepackage{url}
\usepackage{graphics} 
\usepackage{bbm}
\usepackage{wrapfig}
\usepackage{amsmath}
\usepackage{amssymb}
\usepackage{mathtools}
\usepackage{caption}
\usepackage{subcaption}
\usepackage[linesnumbered,ruled,vlined]{algorithm2e}
\usepackage[noadjust]{cite}

\usepackage{booktabs, multirow} 
\usepackage{soul}
\usepackage[table]{xcolor} 
\usepackage{bbding}

\input{macros}

\title{\LARGE \bf Co-learning Planning and Control Policies Constrained by Differentiable Logic Specifications}

\author{
    Zikang Xiong, 
    Daniel Lawson,
    Joe Eappen, 
    Ahmed H. Qureshi, 
    and Suresh Jagannathan
    \thanks{Authors affiliate to Computer Science Department, Purdue University, IN 47906, USA. (E-mail: xiong84@purdue.edu; lawson95@purdue.edu; jeappen@purdue.edu; ahqureshi@purdue.edu; suresh@cs.purdue.edu)}}

\begin{document}

\maketitle
\thispagestyle{empty}
\pagestyle{empty}

\begin{abstract}
    Synthesizing planning and control policies in robotics is a fundamental task, further complicated by factors such as complex logic specifications and high-dimensional robot dynamics.
    This paper presents a novel reinforcement learning approach to solving high-dimensional robot navigation tasks with complex logic specifications by co-learning planning and control policies.
    Notably, this approach significantly reduces the sample complexity in training, allowing us to train high-quality policies with much fewer samples compared to existing reinforcement learning algorithms. In addition, our methodology streamlines complex specification extraction from map images and enables the efficient generation of long-horizon robot motion paths across different map layouts.
    Moreover, our approach also demonstrates capabilities for high-dimensional control and avoiding suboptimal policies via policy alignment.
    The efficacy of our approach is demonstrated through experiments involving simulated high-dimensional quadruped robot dynamics and a real-world differential drive robot (TurtleBot3) under different types of task specifications.
\end{abstract}

\input{sections/1_introduction.tex}
\input{sections/2_related_work.tex}
\input{sections/3_background.tex}
\input{sections/4_approach.tex}

\input{sections/5_experiment.tex}
\input{sections/6_conclusion.tex}

\bibliographystyle{IEEEtran}
\bibliography{refs}

\end{document}

%% file: macros.tex
\newcommand{\X}{\bigcirc}
\newcommand{\E}[2]{\Diamond_{[#1,#2]}}
\newcommand{\G}[2]{\square_{[#1,#2]}}
\newcommand{\U}[2]{\ \mathcal{U}_{[#1,#2]}}
\newcommand{\expctation}{\mathop{\mathbb{E}}}

\definecolor{airforceblue}{rgb}{0.36, 0.54, 0.66}
\definecolor{applegreen}{rgb}{0.55, 0.71, 0.0}
\definecolor{ballblue}{rgb}{0.13, 0.67, 0.8}

\newcommand{\traj}{\tau}

\renewcommand{\implies}{\Rightarrow}
\newcommand{\argmax}{\mathop{\mathrm{argmax}}}

%% file: sections/1_introduction.tex
\section{Introduction}
\label{sec:intro}

\begin{figure*}[ht]
    \centering
    \includegraphics[width=\linewidth]{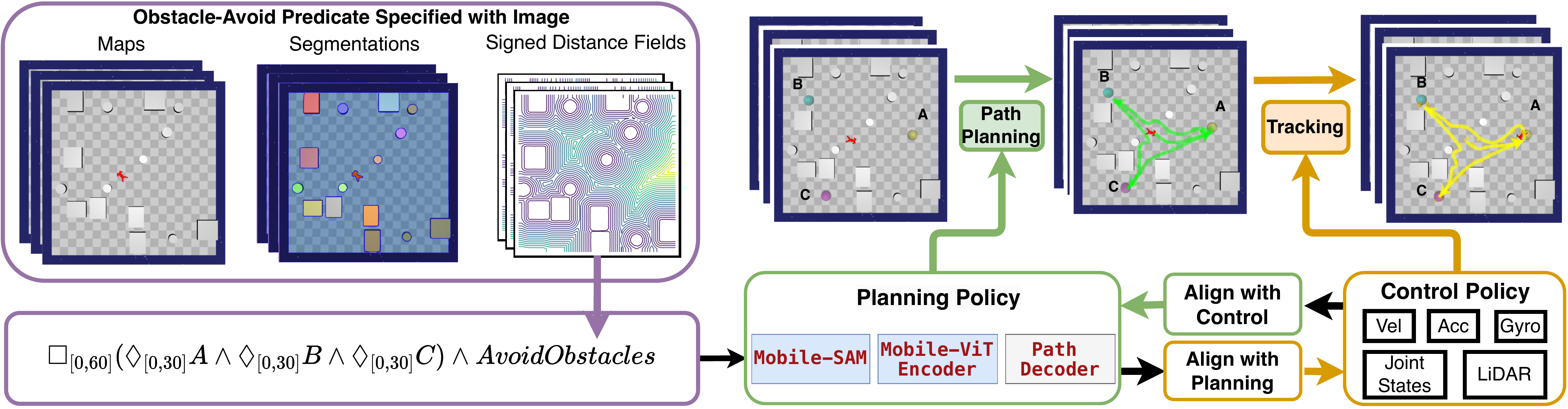}
    \caption{ The \textcolor[rgb]{0.588,0.451,0.651}{differentiable specification} encodes a task visiting location $A, B, C$ repeatedly, and the predicate $AvoidObstacles$ is a Signed Distance Field (SDF) directly specified from a map image.  \emph{Differentiable} symbolic specifications constrain policy updates and notably reduce the \emph{sample complexity} of training policies. Based on a different initial position and map layouts, the \textcolor[rgb]{0.51,0.702,0.4}{planning policy} crafts paths in compliance with the given specification. The \textcolor[rgb]{0.843,0.608,0}{control policy} is a goal-conditioned policy, following the paths laid out by the planner. It learns to meet goals generated by the \textcolor[rgb]{0.51,0.702,0.4}{planning policy}. The \textcolor[rgb]{0.51,0.702,0.4}{planning policy} and \textcolor[rgb]{0.843,0.608,0}{control policy} are aligned with each other during training. }
    \label{fig:intro-example}
    \vspace{-.5cm}
\end{figure*}

Synthesizing planning and control policies is among the core tasks of robotics. The planning policy determines a path comprising a sequence of robot configurations between the given start and goal. In contrast, the control policy helps robots interact with the physical environment allowing them to follow the given plan to reach a goal. However, synthesizing planning and control policies is challenging since the high-dimensional robot dynamics introduce a vast space of possible movements and reactions, making it difficult to predict and control every potential outcome. In addition, the planning policy may be subject to complex logic specifications, which dictate how a robot should behave under certain conditions or constraints. These specifications can range from simple commands like ``reaching A'', and ``avoiding B'' to intricate sets of rules like ``surveilling crucial spots and stopping after a certain condition is triggered''. Merging planning and control while adhering to these logic rules and managing the inherent complexities of robot dynamics presents a significant challenge, even for state-of-the-art approaches.

We leverage Reinforcement Learning (RL) for high-dimensional robot planning and control under complex dynamics and sensor observation.
To ensure that robots adhere to behavioral rules, temporal logic can be incorporated into RL \cite{fu2014probably,li2017reinforcement,hasanbeig2019reinforcement,jothimurugan2019composable,icarte2022reward}. This offers a structured way to define and check a robot's behavior over time, ensuring tasks are completed in order. However, incorporating temporal logic into RL presents its own challenges. For example,  previous approaches~\cite{fu2014probably,li2017reinforcement,hasanbeig2019reinforcement,jothimurugan2019composable,lcrl_tool,icarte2022reward,jiang2020temporal,bozkurt2020control,xu2020joint,zhang2022temporal} that have used temporal logic for reward shaping, involves the construction of complicated reward functions that require large amounts of samples \cite{dulac2019challenges}. Furthermore, handcrafting these temporal logic specifications can be challenging. For instance,  situations where map layouts are represented as images or have irregular obstacle shapes present numerous challenges in defining precise, actionable logical rules. Although existing solvers \cite{kurtz2022mixed} can find robot paths under a given logical specification, they lack scalability and can only handle linear and, to some extent, quadratic constraints, which is not ideal for practical scenarios, as highlighted in our experiments. Furthermore, these methods only account for path planning without considering errors from the underlying controller or vice versa, which often leads to failure in executions.

Therefore, we propose a novel approach, called Differentiable Specifications Constrained Reinforcement Learning (DSCRL), to address the above challenges using the following key features:

\noindent \textbf{Lower Sample Complexity}: We introduce a new methodology that integrates differentiable specifications into constrained RL, which significantly lowers the sample complexity of training.

\noindent \textbf{Control and Planning Alignment}: Learning both planning and control policies separately faces the challenges of suboptimal policies due to lack of alignment. We analyze this issue and demonstrate that our approach allows for better alignment between planning and control policies.

\noindent \textbf{Efficient Planning}: Our approach generates long horizon plans in cluttered environments, which is typically hard for other RL algorithms as well as existing Signal Temporal Logic (STL) solvers to handle.

\noindent \textbf{Specifications from Image}: Our approach can extract irregular obstacle specifications directly from different map images, and solve them efficiently with a novel neural planning policy built upon pre-trained backbones.

\noindent \textbf{High-dimensional Policy}: Our approach has been effectively tested on simulated high-dimensional quadruped robot dynamics and a real-world differential drive robot (TurtleBot3) with LiDAR observation.

%% file: sections/2_related_work.tex
\section{Related Work}
\label{sec:related}
\noindent \textit{Temporal Logic and RL: } This work considers symbolic specifications written in temporal logic \cite{first_ltl,maler2004monitoring}, which are widely used in specifying various planning and control tasks\cite{fainekos2005temporal,cbelta_ltl,kg_ltl,receding_ltl_planning}. Along with the advances of RL in stochastic, model-free settings, a line of work ~\cite{fu2014probably,li2017reinforcement,hasanbeig2019reinforcement,jothimurugan2019composable,lcrl_tool,icarte2022reward,jiang2020temporal,bozkurt2020control,xu2020joint,zhang2022temporal} have considered combining LTL-specified tasks with reinforcement learning. However, these approaches formulate such integration using temporal logic as rewards, leading to inefficient or even intractable algorithms~\cite{yang2021reinforcement}.

\noindent \textit{Constrained RL:} Constrained reinforcement learning \cite{schulman2015trust,achiam2017constrained,schulman2017proximal} provides an efficient approach to enforce constraints on policies. Given a differentiable specification, these methods can enforce constraints only on the policy without shaping rewards. However, prior work did not consider how to synergistically take advantage of policies with differentiable formal constraints \cite{LeungArechigaEtAl2021}.

\noindent \textit{Gradient and Neural Path Planning:} Our work also relates to gradient-based motion planning \cite{LeungArechigaEtAl2021,leung2020back,ratliff2009chomp,campana2016gradient,dawson2022robust}. While these techniques are typically only applicable when the dynamics are known and deterministic, our focus is on planning in the face of unknown and stochastic dynamics. Our high-level planning policy is related to planning networks \cite{qureshi2019motion}. However, learning a planning neural network in an RL loop with task specifications, which aligns both the planning policy and control policy, has not been explored previously.

%% file: sections/3_background.tex
\section{Background}

\subsection{Policy Gradient}
Consider a trajectory \( \tau \) sampled from policy \( \pi_\theta \). The expected return \( R(\tau) \) for a policy parameterized by \( \theta \) is given by
\( J(\theta) = \mathbb{E}_{\tau \sim \pi_\theta}[R(\tau)] \). The policy gradient theorem \cite{williams1992simple} states that the gradient of the expected return is given by
\( \nabla_\theta J(\theta) = \mathbb{E}_{\tau \sim \pi_\theta}[\nabla_\theta \mathcal{P} (\pi_\theta, \tau) R(\tau)]\),
where $\mathcal{P} (\pi_\theta, \tau)$ is the log probability of trajectory $\tau$ under policy $\pi_\theta$. This implies that the probability of generating a trajectory $\tau$ with a high return is maximized compared to those with a lower return. In practice, a baseline $b(\tau)$ is also subtracted from the reward to reduce the variance of the gradient estimator \cite{wu2018variance}.

\subsection{Signal Temporal Logic}

The syntax of STL contains both first-order logic operators $\land$ (and), $\lnot$ (not), $\lor$ (or), $\implies$ (implies), and temporal operators $\X$ (next), $\E{a}{b}$ (eventually), $\G{a}{b}$ (globally), $\U{a}{b}$ (until). The initial time $a$ and end time $b$ ``truncate'' a path. For example, $\G{a}{b}$ qualifies property that globally holds during time $a$ and $b$. The syntax of STL is recursively defined via the following grammar:
\begin{equation}
    \begin{aligned}
        \phi := & \top \mid \bot \mid \mathcal{F} \mid \lnot \phi \mid \phi \land \psi \mid \phi \lor \psi \mid \phi \implies \psi \\
                & \mid \X\phi \mid \E{a}{b} \phi \mid \G{a}{b} \phi \mid \phi \U{a}{b} \psi,
    \end{aligned}
\end{equation}
where $\mathcal{F}: \mathbb{R}^n \rightarrow \mathbb{R}$ is a predicate function mapping a state to a real value (e.g., a function computes the distance to an obstacle surface). The STL semantics $\rho(\traj, \phi): \mathbb{R}^{T\times n} \rightarrow \mathbb{R}$ is defined over a formula $\phi$ and a trajectory $\traj$ with length $T$, and maps them to a real value \cite{LeungArechigaEtAl2021}. For convenience, we denote $\phi(\tau):= \rho(\traj, \phi)$. As an example, the semantics of $\G{a}{b} \phi$ is:
\begin{align}
    \rho(\traj, \G{a}{b} \phi ) = \min_{t \in [a, b]}\rho(\traj_{[t:b]}, \phi),
\end{align}
meaning that the formula $\G{a}{b} \phi$ holds if and only if $\phi$ holds at every time step ($\forall t \in [a, b], \phi(\tau_{[t:b]}) > 0$). 
$\tau_{[t:b]}$ is a path slice from $t$ to $b$. 
The full semantics can be found in previous works such as \cite{LeungArechigaEtAl2021,smoothStl}. 
In practice, $\min, \max$ in STL semantics can cause gradient vanishing, which can be alleviated by using soften techniques \cite{smoothStl}.

%% file: sections/4_approach.tex
\section{Approach}
This section presents the key components of our method for solving complex robot navigation tasks.
In Sec.~\ref{sec:planning-policy}, we will introduce a novel planning policy \(\pi^{h}_{\phi}\) which generates paths based on task specification \(\phi\) and map images, while also aligns with the control policy \(\pi^{l}\). Then, we introduce the control policy \(\pi^{l}\) which follows paths and aligns itself with the planning policy \(\pi^{h}_{\phi}\) in Sec.~\ref{sec:control-policy}.

\subsection{Planning Policy}
\label{sec:planning-policy}

The planning policy $\pi^{h}_{\phi}$ is trained to meet a task specification $\phi$. Given an initial position $x_0$ and an arbitrary map image $I$, the planned path $\tau$ is sampled from the distribution predicted by the planning policy, where $\tau= (g_0, g_1, \dots, g_T) $, $g_0 = x_0$, and $T$ is the max time horizon, i.e.,\(\tau \sim \pi^{h}_\phi(x_0, I)\).
STL has the following definition:
\begin{equation}
    \phi(\tau) =
    \begin{cases}
        > 0, & \text{if } \tau \text{ satisfies } \phi,        \\
        < 0, & \text{if } \tau \text{ does not satisfy } \phi.
    \end{cases}
    \label{eq:stl-true-false}
\end{equation}

\subsubsection{Differentiable STL and Control Feedback}
The quantitative semantics of $\phi$ is provided by STL \cite{maler2004monitoring}. Because $\phi$ is differentiable \cite{LeungArechigaEtAl2021}, we can directly use $\phi$ as a part of the training objective, and turn the policy search into a gradient optimization problem. Meanwhile, to align the planning policy $\pi^h_\phi$ with the feedback from the control policy $\pi^l$, while also satisfying $\phi$, we formulate the problem by minimizing the following loss function:

\begin{align}
    \label{eq:high-level-obj-func}
    \mathcal{L}_{\pi^h_\phi} =  -\expctation_{\substack{I \sim \mathcal{I}, x_0 \sim \mathcal{X}_0, \\ \tau \sim \pi^h_\phi(x_0, I)}} \biggl(\underbrace{\mathcal{P}(\tau, \pi^h_\phi) \cdot r^h(\tau, \pi^l)}_{\text{PG}} + \lambda \phi(\tau) \biggr).
\end{align}
The reward $r^h(\tau, \pi^l)$ here is the negative of the steps to reach the goal $g_T$. When following the path $\tau$ with the control policy $\pi^l$, fewer steps mean a higher reward. The $\mathcal{P}(\tau, \pi^h_\phi)$ is the log probability of the path $\tau$ under policy $\pi^h_\phi$. The task specification $\phi$ is formulated as a Lagrangian constraint with a multiplier $\lambda$. Optimizing ~\eqref{eq:high-level-obj-func} will maximize the expectation of reward, which is positively correlated to the expectation of $\mathcal{P}(\tau, \pi^h_\phi) \cdot r^h(\tau, \pi^l)$, and generate paths satisfied $\phi$ (i.e., $\phi(\tau) > 0$). In practice, $r^h(\tau, \pi^l)$ will subtract a baseline to reduce the policy gradient variance \cite{wu2018variance}. The Lagrangian multiplier $\lambda$ can be dynamically updated for better performance \cite{stooke2020responsive}. 

One may only train the planning policy $\pi^h_\phi$ with the task specification $\phi$ without control feedback (i.e., ignoring $\mathcal{P}(\tau, \pi^h_\phi) \cdot r^h(\tau, \pi^l)$). However, due to the complexity of the high-dimensional dynamics we are handling, it is impractical to encode dynamics constraints in STL. Thus, despite the fact that the planner trained with $\phi(\tau)$ can generate paths satisfying $\phi$, these paths are not necessary to be easy to follow by the control policy. For example, it may generate a path with lots of sharp turns. More importantly, choosing the goal for a high-dimensional, non-linear goal-conditioned control policy $\pi^l$ is non-trivial. Naively choosing a path based on Euclidean distance does not necessarily mean the minimum in the steps. Thus, optimizing the $\pi^h_\phi$ with control feedback is necessary to avoid suboptimal policies.

Previous RL approaches \cite{fu2014probably,li2017reinforcement,hasanbeig2019reinforcement,lcrl_tool,jothimurugan2019composable,jiang2020temporal,bozkurt2020control,xu2020joint,icarte2022reward,zhang2022temporal} have investigated training policies with STL. However, these approaches are based on reward shaping, which is different from our constraint-based formalization. The reward-shaping formulation minimizes the following loss function:
\begin{align}
    \label{eq:reward-shaping-obj-func}
    \mathcal{L}_{RS} =  -\expctation_{\substack{I \sim \mathcal{I}, x_0 \sim \mathcal{X}_0, \\ \tau \sim \pi^h_\phi(x_0, I)}} \underbrace{ \mathcal{P}(\tau, \pi^h_\phi) \cdot \biggl( r^h(\tau, \pi^l) + \lambda \phi(\tau) \biggr)}_{\text{PG}}.
\end{align}
The policy gradient part of \eqref{eq:high-level-obj-func} and \eqref{eq:reward-shaping-obj-func} is marked as PG. Getting an accurate policy gradient typically requires many samples, and a more complex reward function typically requires more samples to estimate the policy gradient. Thus, as illustrated in \eqref{eq:high-level-obj-func}, decoupling the complex STL specification $\phi$ from the policy gradient part and directly backpropagating through $\phi$ is expected to reduce the sample complexity in training (i.e., converge faster). In addition, the estimated policy gradients are only meaningful around existing samples \cite{schulman2015trust,schulman2017proximal}. However, during iterative updates, the policy can quickly shift to a region with few samples and make the policy gradient updates random. In contrast, when backpropagating through $\phi$, as the gradient is not estimated from ``local'' samples, but computed from $\phi$, our objective \eqref{eq:high-level-obj-func} also demonstrates more stable performance during training.

\subsubsection{Model Design of Planning Policy}

\begin{figure}[h]
    \centering
    \includegraphics[width=\linewidth]{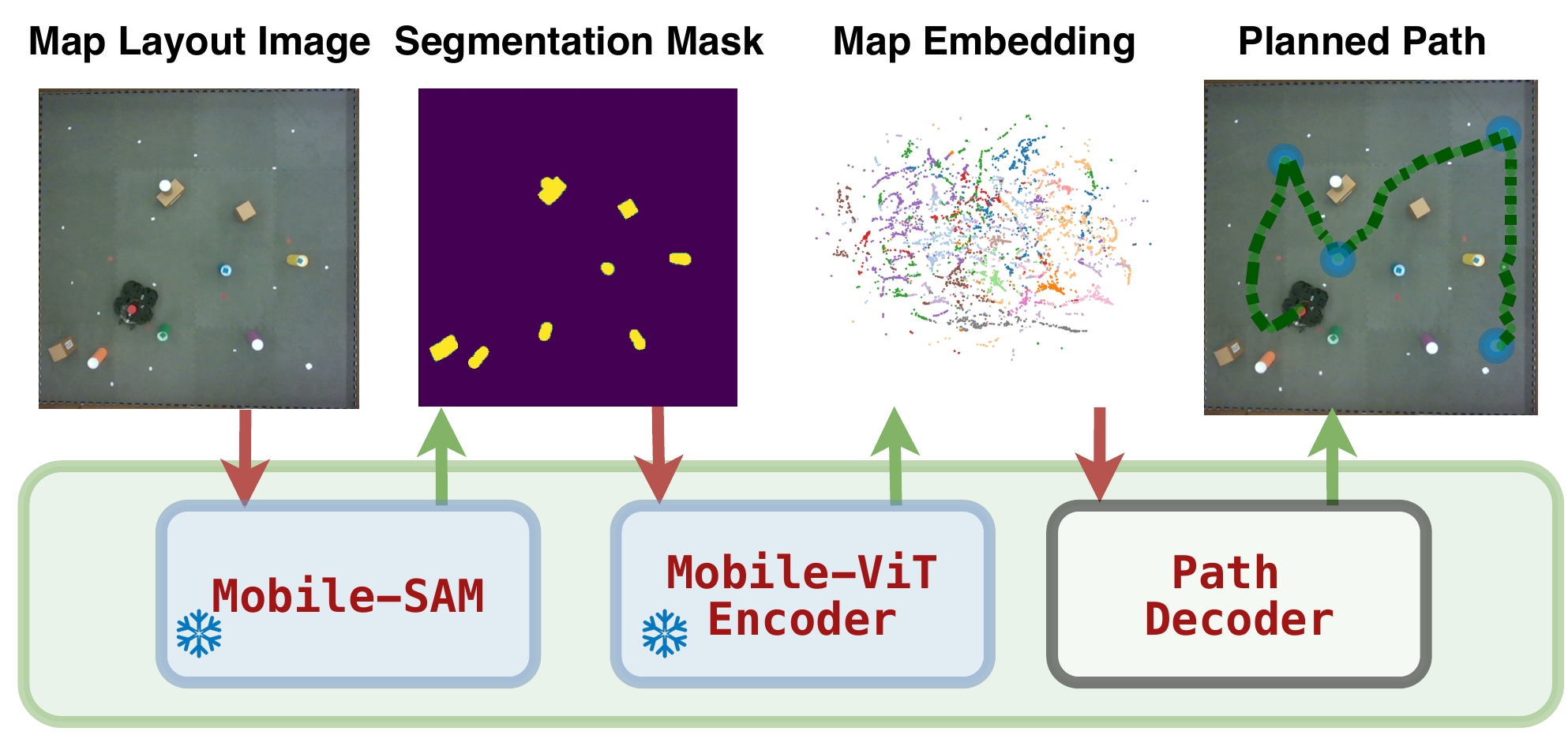}
    \caption{The planning policy model has three parts. The Mobile-SAM \cite{zhang2023faster}, whose input is an arbitrary map image, outputs a mask of obstacles. The mask image is then fed into a Mobile-ViT encoder \cite{mehta2021mobilevit} that outputs the map embedding. The transformer path decoder takes the map embedding and the initial position as input and outputs the deviation between subgoals. Mobile-SAM and Mobile-ViT are pre-trained models with frozen weights  (denoted by \SnowflakeChevron), and the decoder is trained with the loss function in \eqref{eq:high-level-obj-func}.}
\end{figure}

Given an arbitrary map layout image \( I \), the planning policy model processes it in stages to make path planning efficient. Initially, to reduce the noise inherent in real images, the MobileSAM \cite{zhang2023faster} extracts an obstacle mask $M$ from the map image \( I \).
Subsequently, the MobileViT encoder \cite{mehta2021mobilevit} maps this mask image \( M \) into a map embedding $E$.
Lastly,  $Decoder_\theta$,
parameterized by \( \theta \) (the component we exclusively trained) computes subgoal deviations, denoted as \( \frac{d \tau}{d t} \), using the map embedding \( E \) and the robot's initial position \( x_0 \):
\begin{align}
    \label{eq:decoder}
    \frac{d \tau}{d t}  = tanh(Decoder_\theta(E, x_0)) \times \Delta g.
\end{align}
Here, $\Delta g$ is the maximum deviation between subgoals. The deviation is scaled by a $tanh$ function to ensure that the subgoals are within a reasonable range. The subgoal in a planned path $\tau = [g_0, g_1, \dots, g_T]$ is generated by integrating the deviation with the initial position $x_0$:
$
    g_t = x_0 + \int_{0}^{t} \frac{d \tau}{d t} dt.
$
The decoder is a transformer decoder \cite{vaswani2017attention} which will be trained with the loss function in \eqref{eq:high-level-obj-func}.

\subsubsection{Extract Obstacle Specification from Map Image}
\label{sec:extract-obstacle-specification-from-map-image}


One limitation of STL is that all specifications have to be encoded by users, including all obstacles on the map. This can be challenging as the map layout becomes complicated and implicit (e.g., the map is given as an image). In addition, state-of-the-art STL solvers, such as \cite{kurtz2022mixed}, can only encode obstacles with linear or limited quadratic constraints, making it challenging to handle real-world obstacles with irregular shapes.

Our approach extends STL with specifications directly generated from images, by converting them into a Signed Distance Field (SDF) \cite{malladi1995shape}.
The SDF is a function that returns the distance to the nearest obstacle surface for any given point in an obstacle mask $M$:
\begin{align}
    \label{eq:sdf}
    SDF(g, M) = sign(g) \cdot \min_{p \in M} \| g - p \|_2,
\end{align}
Here, $sign(g)$ is negative if $g$ is inside an obstacle, and positive if $g$ is outside an obstacle. We directly use $SDF$ as an obstacle avoidance predicate in STL, which is consistent with the semantics of STL \eqref{eq:stl-true-false}. As an example, the predicate $AvoidObstacles$ in Fig.~\ref{fig:intro-example} is defined as:
\begin{align}
    \label{eq:avoid-obstacle}
    AvoidObstacles = \G{0}{T} SDF (\cdot, M).
\end{align}
The $\G{0}{T}$ is a $\min$ operator defined in the path $\tau = [g_0, g_1, \dots, g_T]$, which returns the minimum value of $SDF(g_t, M)$ for all $g_t$. In other words, $AvoidObstacles(\tau)$ is positive if and only if $\forall g_t \in \tau, SDF(g_t, M) > 0$, which means that all $g_t$ are outside the obstacles.

The mask $M$ is stored as a binary image, where the obstacle pixels are 1 and the free space pixels are 0. We can extract an SDF from the mask image $M$ with a KD-Tree \cite{kdtree}. Given a mask image with $W \times H$ pixels, we build a KD-Tree with the outline points of the obstacles, where each point is a 2D index of a pixel in the mask image. A \( \mathit{KDTree_M} \) constructed specifically for a mask \( M \), when input a pixel index, returns the nearest pixel that is part of an obstacle outline within the mask \( M \). If the field size is $m \times n$ meters in the real world, we can compute a transformation $\mathcal{T}$ from the pixel index to the real-world coordinate, and its inversion $\mathcal{T}^{-1}$, with $W, H, m, n$. An SDF is implemented as:
\begin{align}
    \label{eq:sdf-kd-tree}
    SDF(g, M) = sign(g) \|\mathcal{T} \cdot \mathit{KDTree_M}(\mathcal{T}^{-1} \cdot g) - g \|_2.
\end{align}
The gradient can be backpropagated through \eqref{eq:sdf-kd-tree} to update the planning policy.

\subsection{Control Policy}
\label{sec:control-policy}

The low-level control policy $\pi^l(o_t \mid g)$ is a goal-conditioned policy \cite{schaul2015universal} trained with PPO~\cite{schulman2017proximal}, and $o_t$ is the robot observation at time $t$. The control reward $r^l_t$ at time $t$ is defined as
\begin{align}
    r^l(o_t, o_{t+1}, g) = \| g - pos(o_{t}) \|_2 - \| g - pos(o_{t+1}) \|_2,
\end{align}
where $pos(o_{t})$ and $pos(o_{t+1})$ are the positions of the robot at time $t$ and $t+1$, respectively. The reward is positive if and only if the robot gets closer to the planned goal $g$.

The planning policy $\pi^h_\phi$ generates the goal distribution $\mathcal{G}$ for the control policy. As the planning policy is updated, $\mathcal{G}$ will change. Thus, the control policy $\pi^l$ needs to be updated to follow the new goal distribution. Formally, the general goal for the goal-conditioned RL is to find an optimal policy $\pi^*$ with
\begin{align}
    \label{eq:rl-obj-func}
    \pi^* = \argmax_{\pi} \expctation_{\substack{g \sim \mathcal{G}}} \expctation_{\substack{o_0 \sim \mathcal{O}_0,  a_t \sim \pi(o_t \mid g), \\ o_{t+1} \sim P(o_{t+1} \mid o_t, a_t)}} \biggl( \sum_{t=0}^{T} \gamma^t r^l(o_t, o_{t+1}, g) \biggr),
\end{align}
where $\mathcal{O}_0$ is the initial observation distribution, and $P(o_{t+1} \mid o_t, a_t)$ is the transition function. The $\gamma$ is the discount factor.
Let $V^{\pi}(g) = \expctation_{\substack{o_0 \sim \mathcal{O}_0,  a_t \sim \pi(o_t \mid g), \\ o_{t+1} \sim P(o_{t+1} \mid o_t, a_t)}} \biggl( \sum_{t=0}^{T} \gamma^t r^l(o_t, o_{t+1}, g) \biggr)$ be the value function of the control policy $\pi$ with goal $g$, the objective function in \eqref{eq:rl-obj-func} can be rewritten as:
\begin{align}
    \label{eq:rl-obj-func-2}
    \pi^* = \argmax_{\pi} \sum_g P_{\mathcal{G}}(g) V^{\pi}(g).
\end{align}
The $P_{\mathcal{G}}(g)$ is the probability of the goal $g$ under the goal distribution $\mathcal{G}$. The change of the goal distribution $\mathcal{G}$ (as planning policy updates) will change $P_{\mathcal{G}}$, and thus the optimal policy $\pi^*$ shifts. Thus, the control policy $\pi^l$ needs to be updated for the new goal distribution. This can be achieved by training the control policy with any RL algorithm by sampling the goal $g$ with the new goal distribution \cite{schaul2015universal}.

\subsection{Overall Pipeline}
We summarize the training and deployment pipeline of our approach in this section. First, users express their goals without delving into intricate maps, as exemplified by \( \square_{[0,60]} (\Diamond_{[0,30]} A \land \Diamond_{[0,30]} B \land \Diamond_{[0,30]} C) \) in Fig.~\ref{fig:intro-example}. Next, the $AvoidObstacles$ specification is derived from map images with the method discussed in Sec.~\ref{sec:extract-obstacle-specification-from-map-image}.
For policy training, we initially train \(\pi^l\) using goals sampled from \(\pi^h_\phi\) while keeping \(\pi^h_\phi\) fixed. Then, we train the planning policy \(\pi^h_\phi\) using \eqref{eq:high-level-obj-func} with the control policy \(\pi^l\) fixed. This back-and-forth is important as simultaneous updates can introduce instability due to the evolving objectives in \eqref{eq:high-level-obj-func} and \eqref{eq:rl-obj-func-2}. We iterate through this process until both policies converge. During deployment, the planning policy $\pi^h_\phi$ generates a path $\tau$ from the initial position $x_0$ and an arbitrary map image $I$. The control policy $\pi^l$ follows $\tau$ to achieve the task $\phi$.

%% file: sections/5_experiment.tex
\begin{figure*}[ht]
    \centering
    \includegraphics[width=\linewidth]{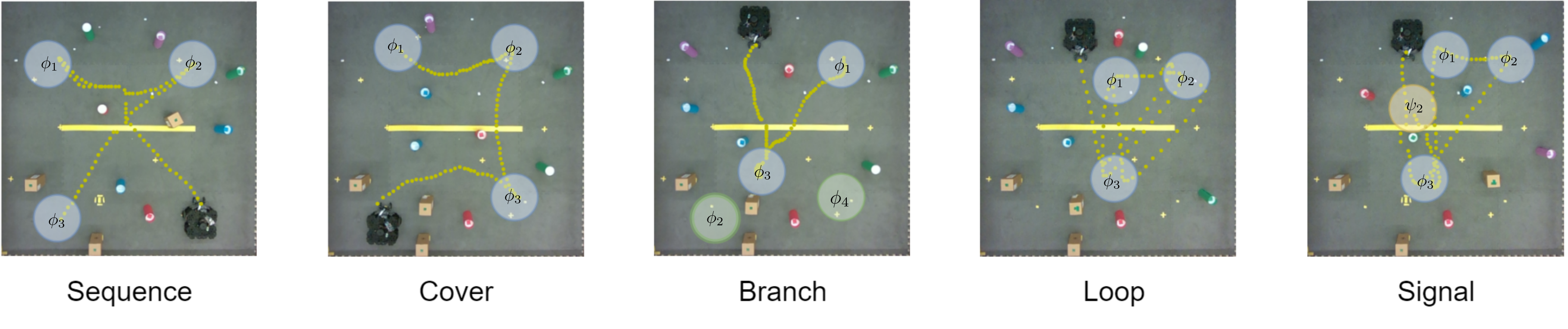}
    \vspace{-.75cm}
    \caption{In the \textit{Sequence} task, the \textit{Turtlebot3} must visit the three blue regions sequentially. The \textit{Cover} task requires the robot to traverse all three blue regions in any sequence. For the \textit{Branch} task, the robot chooses between visiting either the two blue regions or the two green regions. The \textit{Loop} task asks the robot to cyclically traverse the three blue regions. Lastly, the \textit{Signal} task requires the robot to loop among three blue regions and exit the loop after visiting the lower blue region twice, and finally visiting the yellow region. The yellow tape in the center is used for calibrating the coordinates. }
    \label{fig:tasks}
    \vspace{-0.5cm}
\end{figure*}

\section{Experiments}
\label{sec:exp}

\subsection{Setup}
\subsubsection{Tasks}
The tasks evaluated are depicted in Fig.~\ref{fig:tasks}.
The STL specifications of the \textit{Sequence} task is $\bigwedge_{i=1}^{N} \E{t_{i}}{t_{i+1}} \phi_{i}$,
where $\phi_i$ is a predicate that corresponds to reaching each blue region. The robot must visit the region specified by $\phi_1, \phi_2, \cdots, \phi_{N}$ in sequence.
The \textit{Cover} task is specified by $\bigwedge_{i=1}^{N} \E{0}{T} \phi_i$. Here the robot is asked to cover all the specified regions in any given order.
The \textit{Branch} task is specified by $\bigvee_{i=1}^{\frac{N}{2}} (\E{0}{T} \phi_i \land \E{0}{T} \phi_{\frac{N}{2} + i})$, the robot must either visit the branch of blue regions ($\phi_1$, $\phi_3$) or the branch of green regions ($\phi_2$, $\phi_4$).
The \textit{Loop} task $\G{0}{T - \lfloor\frac{T}{M}\rfloor} (\bigwedge_{i=1}^{N} \E{0}{\lfloor\frac{T}{M}\rfloor}\phi_i)$ asks the robot to cyclically traverse the blue regions specified by $\phi_i$ for $M$ times.
Lastly, the \textit{Signal} task is specified by $\G{0}{T - \lfloor\frac{T}{M}\rfloor} (\bigwedge_{i=1}^{N} \E{0}{\lfloor\frac{T}{M}\rfloor}\phi_i) \U{0}{1} \psi_{1} \land \E{0}{T} \psi_{2}$.
The robot should loop among the three blue regions ($\phi_i$) until it visits the lower blue region twice ($\psi_{1}$) and finally visits the yellow region ($\psi_{2}$). All these specifications are appended with $\land \G{0}{T} SDF (\cdot, M)$ extracted from map $M$.

\subsubsection{Observation, Action, and Dynamics}
The \textit{Doggo} (shown in Fig.~\ref{fig:intro-example}) features a 74-dimensional observation space: it has joint states (44D), IMU data (12D), position (2D), and a 16-beam LiDAR for obstacle detection. Its action space is 12-dimensional, comprising 8 hip and 4 ankle control commands. On the other hand, \textit{TurtleBot3} (shown in Fig.~\ref{fig:tasks}) utilizes 4-dimensional IMU data (including x-axis position, y-axis position, and the sin and cos of rotation) and a 36-dimensional down-sampled LiDAR for measuring obstacle distances. Its action commands are dedicated to controlling the robot's linear and angular velocity.

\subsection{Policy Performance and Alignment}

\subsubsection{Metrics}
We evaluate the planning and control policies using  Success Rate (SR) and Time to Reach (TtR) metrics. SR quantifies the proportion of successful plans. A plan is considered successful if the path $\tau$ satisfies the condition ($\phi(\tau) > 0$) without any collisions when followed by the control policy. TtR calculates the duration (in seconds) needed to achieve the final destination. Each learned policy is evaluated on 1000 episodes with different map layouts. There are 10 obstacles for \textit{TurtleBot3} with $2.42 \times 2.42$ meters maps and 15 obstacles for \textit{Doggo} with $3 \times 3$ meters maps.

\begin{table}[!htp]\centering
    \caption{Alignment Experiments. The Success Rate (SR) and Time to Reach (TtR) of the control policy are evaluated on 1000 episodes with randomly sampled map layouts. The performance of aligned policies is reported in the column ``w.'', while the unaligned policies is reported in the column ``w.o.''. The aligned policies perform significantly better than the unaligned policies. All the aligned policies have SR above 96\%.}
    \label{tab: alignment-experiments}
    \scriptsize
    \setlength{\tabcolsep}{4.5pt}
    \begin{tabular}{l|cc|cc|cc|ccc}\toprule
        Robot   & \multicolumn{4}{c|}{\textit{Doggo}}          & \multicolumn{4}{c}{\textit{Turtlebot3}}                                                                                                                                                                                     \\\cmidrule{1-9}
        Metrics & \multicolumn{2}{c|}{SR $\uparrow$} & \multicolumn{2}{c|}{TtR $\downarrow$} & \multicolumn{2}{c|}{SR $\uparrow$} & \multicolumn{2}{c}{TtR $\downarrow$}                                                                                                  \\\cmidrule{1-9}
        Align   & \cellcolor[HTML]{A8A8A8}w.o.       & w.                                    & \cellcolor[HTML]{A8A8A8}w.o.       & w.                                   & \cellcolor[HTML]{A8A8A8}w.o. & w.              & \cellcolor[HTML]{A8A8A8}w.o. & w.             \\\midrule
        \textit{Seq}     & 87.1\%                             & \textbf{97.7}\%                       & 64.6                               & \textbf{46.3}                        & 79.1\%                       & \textbf{98.2\%} & 105.6                        & \textbf{92.7}  \\
        \textit{Cover}   & 83.9\%                             & \textbf{96.4}\%                       & 73.7                               & \textbf{57.0}                        & 73.1\%                       & \textbf{96.1\%} & 92.5                         & \textbf{81.9}  \\
        \textit{Branch}  & 87.2\%                             & \textbf{98.2}\%                       & 56.4                               & \textbf{44.6}                        & 91.1\%                       & \textbf{97.2\%} & 81.4                         & \textbf{73.8}  \\
        \textit{Loop}    & 83.1\%                             & \textbf{98.5}\%                       & 70.6                               & \textbf{51.3}                        & 85.1\%                       & \textbf{98.2\%} & 114.3                        & \textbf{94.5}  \\
        \textit{Signal}  & 79.9\%                             & \textbf{97.1}\%                       & 72.1                               & \textbf{57.3}                        & 79.6\%                       & \textbf{97.8\%} & 118.1                        & \textbf{106.5} \\
        \bottomrule
    \end{tabular}
\end{table}

\subsubsection{Performance and Alignment}
Experiments in Table~\ref{tab: alignment-experiments} show the performance of our final trained policies and the effect of policy alignment. The unaligned planning policy \(\pi^h_\phi\) is trained with STL objective \(\phi\) without control feedback, and the unaligned control policy is trained on random goals. The aligned policies are trained with our approach. All the aligned policies have SR above 96\%, which is significantly higher than the unaligned policies. The aligned policies also have lower TtR than the unaligned policies. The results demonstrate that our approach can work effectively on high-dimensional control tasks with various specifications and maps with high SR and low TtR. Moreover, the results also show that the alignment between planning and control policies is essential for good performance, justifying our algorithm design.

\subsubsection{Policy Performance In A Real TurtleBot3}

\begin{table}[!htp]\centering
    \caption{Deploy trained policy to real TurtleBot3. The first row shows the specs. The second and third rows are identical to TABLE~\ref{tab: alignment-experiments}. The results are computed over 6 sampled episodes each.
    }
    \label{tab: real-world-experiments}
    \scriptsize
    \setlength{\tabcolsep}{4.8pt}
    \begin{tabular}{l|cc|cc|cc|ccc}\toprule
        Spec    & \multicolumn{4}{c|}{\textit{Seq}}           & \multicolumn{4}{c}{\textit{Signal}}                                                                                                                                                                                       \\\cmidrule{1-9}
        Metrics & \multicolumn{2}{c|}{SR$\uparrow$} & \multicolumn{2}{c|}{TtR $\downarrow$} & \multicolumn{2}{c|}{SR $\uparrow$} & \multicolumn{2}{c}{TtR $\downarrow$}                                                                                                \\\cmidrule{1-9}
        Align & \cellcolor[HTML]{A8A8A8}w.o.      & w.                                    & \cellcolor[HTML]{A8A8A8}w.o.       & w.                                   & \cellcolor[HTML]{A8A8A8}w.o. & w.            & \cellcolor[HTML]{A8A8A8}w.o. & w.             \\\midrule
        Stat.     & 3/6                              & \textbf{5}/6                        & 142.7                              & \textbf{108.3}                        & 1/6                         & \textbf{6}/6 & 144.7                     & \textbf{109.1} \\
        \bottomrule
    \end{tabular}
    \vspace{-0.3cm}
\end{table}
We evaluated our policies with and without alignment on a real Turtlebot3 robot for the \textit{Sequence} and \textit{Signal} tasks. 
The results in Table~\ref{tab: real-world-experiments} show that our policies can work effectively in the real world with high SR. 
Compared with the unaligned policies, the aligned policies have higher SR and lower TtR.

\subsection{Sample Efficiency and Comparison with Reward Shaping}

We focused on the number of samples required for our policies to reach convergence. Each sample represents a transition, moving from one observation \(o_t\) to another \(o_{t+1}\) upon taking an action \(a_t\) and acquiring the corresponding reward \(r^l_t\). The high-level reward \(r^h\) is determined after we collect a full trajectory \(\tau\).

To evaluate the efficacy of our methodology, we compare it with the Reward Shaping (RS) strategy \cite{li2017reinforcement} illustrated in \eqref{eq:reward-shaping-obj-func}. Additionally, we benchmarked against the Reward Machine (RM) \cite{icarte2022reward} strategy that shapes rewards with the progression of a pre-defined planned path. For example, given a planned path $A, B, C$, if the planning policy did not reach $A$, the reward is shaped to encourage the planning policy to reach $A$; if the planning policy reached $A$ but did not reach $B$, the reward is shaped to encourage the planning policy to reach $B$, and so on.
Notice that the RM strategy requires an additional solver to generate the planned path, which is a stronger requirement than ours.
Our experimentation spanned two distinct settings:

\subsubsection{Two-Stage Training}
\begin{figure}[h]
    \includegraphics[width=\linewidth]{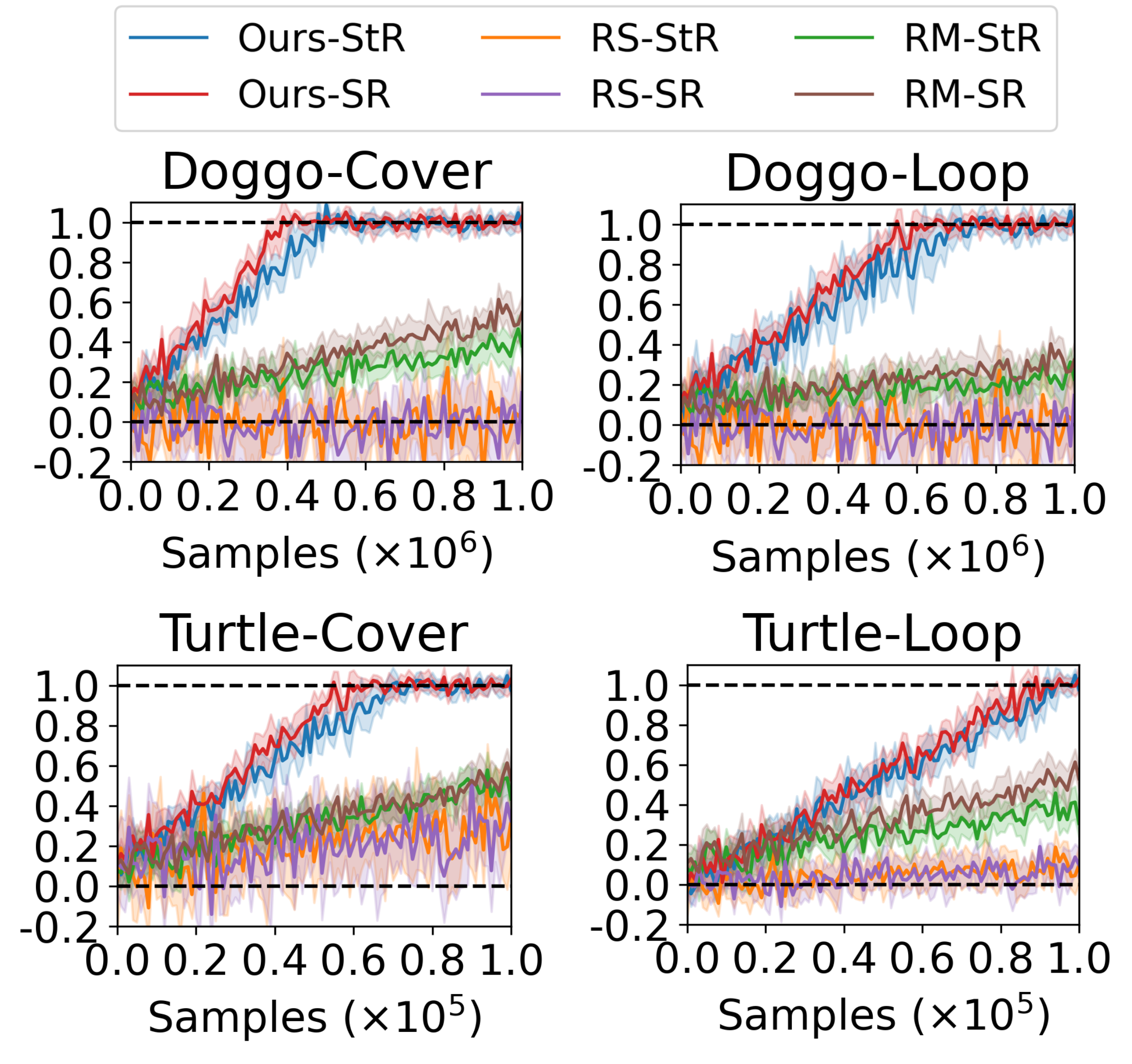}
    \caption{Align pretrained policies on \textit{Cover} and \textit{Loop} tasks. The x-axis is the number of samples, and the y-axis is the normalized score. The upper black dashed line is the score of the aligned policies (``w.'' columns in Table~\ref{tab: alignment-experiments}), and the lower black dashed line is the score of the unaligned policies (``w.o.'' columns in Table~\ref{tab: alignment-experiments}). The values are computed over 5 different random seeds. The results show that our approach can converge with fewer samples than RS and RM.}
    \vspace{-0.3cm}
    \label{fig:align-pretrained-policies}
\end{figure}

In this setup, we first train the planning policy \(\pi^h_\phi\) with STL objective \(\phi\) without control feedback, while the control policy is trained on random goals. Later, in our approach, \(\pi^h_\phi\) was finetuned using \eqref{eq:high-level-obj-func}.
In contrast, RS trains the \(\pi^h_\phi\) with \eqref{eq:reward-shaping-obj-func}, while RM trains \(\pi^h_\phi\) with rewards crafted from a path saftisfying $\phi$. 
Following this, for all methods, \(\pi^l\) is trained on goals sampled from \(\pi^h_\phi\). 
Unlike RS and RM, our method decouples the STL objective from the reward and converges faster (Fig.~\ref{fig:align-pretrained-policies}). The trade-off between sample efficiency and reward function complexity is also evident in RS and RM, with RM outperforming RS with simplified reward functions.

\begin{table}[!htp]\centering
    \caption{Converging samples when training from scratch. The data shows the number of samples required for each algorithm to converge to the scores near Table~\ref{tab: alignment-experiments} ($\pm 5\%$). 
    The data is scaled by the coefficient $c$ next to the robot name.
    A dash ``-'' means that it does not converge in $100 \times c$ samples.
    The results are computed over 5  random seeds.
    }
    \label{tab:scrath-sample-nums}
    \scriptsize
    \setlength{\tabcolsep}{4.5pt}
    \begin{tabular}{l|ccc|cccc}\toprule
        Robot & \multicolumn{3}{c|}{\textit{Doggo} ($\times 10^7$)} & \multicolumn{3}{c}{\textit{Turtlebot3} ($\times 10^6$)}                                                                              \\\cmidrule{1-7}
        Algo  & Ours                               & RS                                      & RM             & Ours                   & RS              & RM             \\\midrule
        \textit{Cover} & \textbf{5.1 $\pm$ 1.2}             & -                                       & 27.4 $\pm$ 7.1 & \textbf{7.9 $\pm$ 1.9} & 69.2 $\pm$ 17.1 & 30.9 $\pm$ 5.4 \\
        \textit{Loop}  & \textbf{5.7 $\pm$ 0.9}             & -                                       & -              & \textbf{9.1 $\pm$ 2.8} & -               & 48.7 $\pm$ 4.9 \\
        \bottomrule
    \end{tabular}
    \vspace{-0.3cm}
\end{table}

\subsubsection{End-to-End Training}
Alternatively, we conduct evaluations by training from scratch where
both \(\pi^h_\phi\) and \(\pi^l\) were initialized with random weights. In Table~\ref{tab:scrath-sample-nums}, we observe that our approach consistently requires fewer samples to converge when compared to the other two approaches. The results also show RM performs better than RS, which is consistent with the results in Fig.~\ref{fig:align-pretrained-policies}.

\subsection{More Scalable STL Planning}
One alternative approach for our problem is solving the STL specifications with an STL solver (such as STLPY \cite{kurtz2022mixed}) and tracking the solved path with a goal-conditioned control policy.
However, this approach has its limitations. First, manual obstacle specification is limited to linear or quadratic constraints. 
Second, paths produced by STLPY, while sound and complete w.r.t specifications, can still lead to collisions or stalls when tracking with an unaligned control policy. In the \textit{Loop} task with 100 timesteps, STLPY paths succeed 14 out of 20 times for \textit{Doggo} and 13 out of 20 times for \textit{TurtleBot3} with the optimal robustness parameter \cite{kurtz2022mixed}. The planning policy's SR is 98.5\% for \textit{Doggo} and 98.2\% for \textit{TurtleBot3} on \textit{Loop} task.
\begin{figure}[h]
    \includegraphics[width=\linewidth]{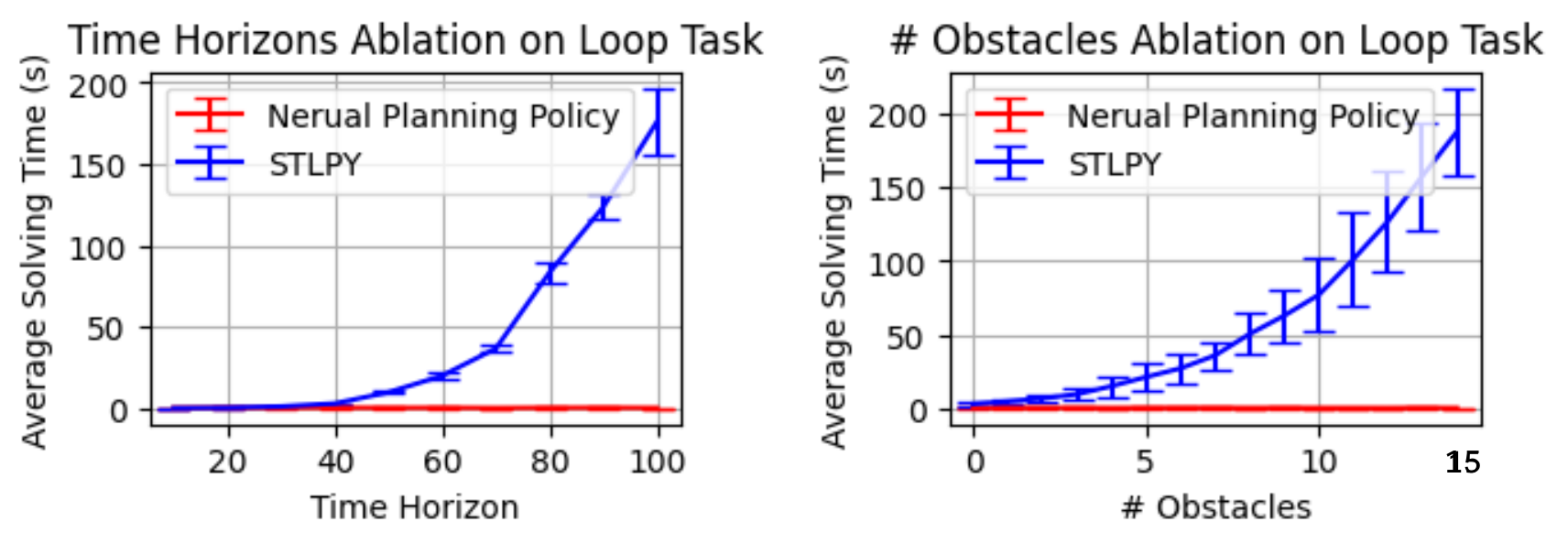}
    \caption{ Planning time against time horizon and obstacle count. The spec for the left is $\G{0}{T - \lfloor\frac{T}{3}\rfloor} (\bigwedge_{i=1}^{3} \E{0}{\lfloor\frac{T}{3}\rfloor}\phi_i)$, where $T$ is the time horizon. The spec for the right is $\G{0}{27} (\bigwedge_{i=1}^{3} \E{0}{13}\phi_i) \land \bigwedge_{i=1}^n Obs_i$, where $\bigwedge_i Obs_i$ represents the rectangle obstacle predicates with linear constraints, and $n$ is the number of obstacles. Results from 20 runs per data point show STLPY's solving time grows exponentially with horizon and obstacles, but the neural planning policy remains consistently fast ($< 0.5$ secs). }
    \vspace{-0.3cm}
    \label{fig:stlpy-scalability}
\end{figure}
Third, as noted by \cite{kurtz2022mixed}, the STL solver has scalability issues shown in Fig.~\ref{fig:stlpy-scalability}. The planning time for the STL solver increases exponentially with the time horizon and the number of obstacles. 

%% file: sections/6_conclusion.tex
\section{Conclusions}
In this paper, we present a new method combining differentiable specifications with constrained RL, which reduces training sample complexity and aligns planning and control policies more efficiently than current reward-shaping techniques. Our method performs well in long-horizon planning in cluttered environments, extracts complex obstacle details from images, and uses them efficiently with our neural planning policy. It also demonstrates advantages in high-dimensional dynamics.

%% file: main.bbl
\begin{thebibliography}{57}
\providecommand{\natexlab}[1]{#1}
\providecommand{\url}[1]{\texttt{#1}}
\expandafter\ifx\csname urlstyle\endcsname\relax
  \providecommand{\doi}[1]{doi: #1}\else
  \providecommand{\doi}{doi: \begingroup \urlstyle{rm}\Url}\fi

\bibitem[Abbeel \& Ng(2004)Abbeel and Ng]{abbeel2004apprenticeship}
Pieter Abbeel and Andrew~Y Ng.
\newblock Apprenticeship learning via inverse reinforcement learning.
\newblock In \emph{Proceedings of the twenty-first international conference on
  Machine learning}, pp.\ ~1, 2004.

\bibitem[Achiam \& Amodei(2019)Achiam and Amodei]{Achiam2019BenchmarkingSE}
Joshua Achiam and Dario Amodei.
\newblock Benchmarking safe exploration in deep reinforcement learning.
\newblock In \emph{Online}, 2019.

\bibitem[Achiam et~al.(2017)Achiam, Held, Tamar, and
  Abbeel]{achiam2017constrained}
Joshua Achiam, David Held, Aviv Tamar, and Pieter Abbeel.
\newblock Constrained policy optimization.
\newblock In \emph{International conference on machine learning}, pp.\  22--31.
  PMLR, 2017.

\bibitem[Andre \& Russell(2002)Andre and Russell]{andre2002state}
David Andre and Stuart~J Russell.
\newblock State abstraction for programmable reinforcement learning agents.
\newblock In \emph{Aaai/iaai}, pp.\  119--125, 2002.

\bibitem[Badreddine et~al.(2022)Badreddine, {d'Avila Garcez}, Serafini, and
  Spranger]{ltn}
Samy Badreddine, Artur {d'Avila Garcez}, Luciano Serafini, and Michael
  Spranger.
\newblock Logic tensor networks.
\newblock \emph{Artificial Intelligence}, 303:\penalty0 103649, 2022.
\newblock ISSN 0004-3702.
\newblock \doi{https://doi.org/10.1016/j.artint.2021.103649}.
\newblock URL
  \url{https://www.sciencedirect.com/science/article/pii/S0004370221002009}.

\bibitem[Bozkurt et~al.(2020)Bozkurt, Wang, Zavlanos, and
  Pajic]{bozkurt2020control}
Alper~Kamil Bozkurt, Yu~Wang, Michael~M Zavlanos, and Miroslav Pajic.
\newblock Control synthesis from linear temporal logic specifications using
  model-free reinforcement learning.
\newblock In \emph{2020 IEEE International Conference on Robotics and
  Automation (ICRA)}, pp.\  10349--10355. IEEE, 2020.

\bibitem[Campana et~al.(2016)Campana, Lamiraux, and
  Laumond]{campana2016gradient}
Myl{\`e}ne Campana, Florent Lamiraux, and Jean-Paul Laumond.
\newblock A gradient-based path optimization method for motion planning.
\newblock \emph{Advanced Robotics}, 30\penalty0 (17-18):\penalty0 1126--1144,
  2016.

\bibitem[Chaudhuri et~al.(2021)Chaudhuri, Ellis, Polozov, Singh,
  Solar{-}Lezama, and Yue]{CE+21}
Swarat Chaudhuri, Kevin Ellis, Oleksandr Polozov, Rishabh Singh, Armando
  Solar{-}Lezama, and Yisong Yue.
\newblock Neurosymbolic programming.
\newblock \emph{Found. Trends Program. Lang.}, 7\penalty0 (3):\penalty0
  158--243, 2021.
\newblock \doi{10.1561/2500000049}.
\newblock URL \url{https://doi.org/10.1561/2500000049}.

\bibitem[Chen et~al.(2018)Chen, Rubanova, Bettencourt, and
  Duvenaud]{neural_ode}
Ricky~TQ Chen, Yulia Rubanova, Jesse Bettencourt, and David~K Duvenaud.
\newblock Neural ordinary differential equations.
\newblock \emph{Advances in neural information processing systems}, 31, 2018.

\bibitem[Cho et~al.(2014)Cho, van Merri{\"e}nboer, Gulcehre, Bahdanau,
  Bougares, Schwenk, and Bengio]{cho-etal-2014-learning}
Kyunghyun Cho, Bart van Merri{\"e}nboer, Caglar Gulcehre, Dzmitry Bahdanau,
  Fethi Bougares, Holger Schwenk, and Yoshua Bengio.
\newblock Learning phrase representations using {RNN} encoder{--}decoder for
  statistical machine translation.
\newblock In \emph{Proceedings of the 2014 Conference on Empirical Methods in
  Natural Language Processing ({EMNLP})}, pp.\  1724--1734, Doha, Qatar,
  October 2014. Association for Computational Linguistics.
\newblock \doi{10.3115/v1/D14-1179}.
\newblock URL \url{https://aclanthology.org/D14-1179}.

\bibitem[Dawson \& Fan(2022)Dawson and Fan]{dawson2022robust}
Charles Dawson and Chuchu Fan.
\newblock Robust counterexample-guided optimization for planning from
  differentiable temporal logic.
\newblock \emph{arXiv preprint arXiv:2203.02038}, 2022.

\bibitem[Fainekos et~al.(2005)Fainekos, Kress-Gazit, and
  Pappas]{fainekos2005temporal}
Georgios~E Fainekos, Hadas Kress-Gazit, and George~J Pappas.
\newblock Temporal logic motion planning for mobile robots.
\newblock In \emph{Proceedings of the 2005 IEEE International Conference on
  Robotics and Automation}, pp.\  2020--2025. IEEE, 2005.

\bibitem[Florensa et~al.(2017)Florensa, Duan, and
  Abbeel]{florensa2017stochastic}
Carlos Florensa, Yan Duan, and Pieter Abbeel.
\newblock Stochastic neural networks for hierarchical reinforcement learning.
\newblock In \emph{International Conference on Learning Representations}, 2017.
\newblock URL \url{https://openreview.net/forum?id=B1oK8aoxe}.

\bibitem[Fu \& Topcu(2014)Fu and Topcu]{fu2014probably}
Jie Fu and Ufuk Topcu.
\newblock Probably approximately correct mdp learning and control with temporal
  logic constraints.
\newblock \emph{arXiv preprint arXiv:1404.7073}, 2014.

\bibitem[Gilpin et~al.(2021)Gilpin, Kurtz, and Lin]{smoothStl}
Yann Gilpin, Vince Kurtz, and Hai Lin.
\newblock A smooth robustness measure of signal temporal logic for symbolic
  control.
\newblock \emph{IEEE Control Systems Letters}, 5\penalty0 (1):\penalty0
  241--246, 2021.
\newblock \doi{10.1109/LCSYS.2020.3001875}.

\bibitem[Hasanbeig et~al.(2018)Hasanbeig, Abate, and
  Kroening]{hasanbeig2018logically}
Mohammadhosein Hasanbeig, Alessandro Abate, and Daniel Kroening.
\newblock Logically-constrained reinforcement learning.
\newblock \emph{arXiv preprint arXiv:1801.08099}, 2018.

\bibitem[Hasanbeig et~al.(2019{\natexlab{a}})Hasanbeig, Abate, and
  Kroening]{hasanbeig2019certified}
Mohammadhosein Hasanbeig, Alessandro Abate, and Daniel Kroening.
\newblock Certified reinforcement learning with logic guidance.
\newblock \emph{arXiv preprint arXiv:1902.00778}, 2019{\natexlab{a}}.

\bibitem[Hasanbeig et~al.(2019{\natexlab{b}})Hasanbeig, Kantaros, Abate,
  Kroening, Pappas, and Lee]{hasanbeig2019reinforcement}
Mohammadhosein Hasanbeig, Yiannis Kantaros, Alessandro Abate, Daniel Kroening,
  George~J Pappas, and Insup Lee.
\newblock Reinforcement learning for temporal logic control synthesis with
  probabilistic satisfaction guarantees.
\newblock In \emph{2019 IEEE 58th Conference on Decision and Control (CDC)},
  pp.\  5338--5343. IEEE, 2019{\natexlab{b}}.

\bibitem[Hasanbeig et~al.(2020)Hasanbeig, Abate, and
  Kroening]{hasanbeig2020cautious}
Mohammadhosein Hasanbeig, Alessandro Abate, and Daniel Kroening.
\newblock Cautious reinforcement learning with logical constraints.
\newblock \emph{arXiv preprint arXiv:2002.12156}, 2020.

\bibitem[Hasanbeig et~al.(2022)Hasanbeig, Kroening, and Abate]{lcrl_tool}
Mohammadhosein Hasanbeig, Daniel Kroening, and Alessandro Abate.
\newblock {LCRL}: Certified policy synthesis via logically-constrained
  reinforcement learning.
\newblock In \emph{International Conference on Quantitative Evaluation of
  SysTems}. Springer, 2022.

\bibitem[He et~al.(2016)He, Zhang, Ren, and Sun]{he2016deep}
Kaiming He, Xiangyu Zhang, Shaoqing Ren, and Jian Sun.
\newblock Deep residual learning for image recognition.
\newblock In \emph{Proceedings of the IEEE conference on computer vision and
  pattern recognition}, pp.\  770--778, 2016.

\bibitem[Icarte et~al.(2022)Icarte, Klassen, Valenzano, and
  McIlraith]{icarte2022reward}
Rodrigo~Toro Icarte, Toryn~Q Klassen, Richard Valenzano, and Sheila~A
  McIlraith.
\newblock Reward machines: Exploiting reward function structure in
  reinforcement learning.
\newblock \emph{Journal of Artificial Intelligence Research}, 73:\penalty0
  173--208, 2022.

\bibitem[Jansen et~al.(2020)Jansen, K{\"o}nighofer, Junges, Serban, and
  Bloem]{jansen2020safe}
Nils Jansen, Bettina K{\"o}nighofer, Sebastian Junges, AC~Serban, and Roderick
  Bloem.
\newblock Safe reinforcement learning using probabilistic shields.
\newblock 2020.

\bibitem[Jha et~al.(2019)Jha, Tiwari, Seshia, Sahai, and Shankar]{jha2019telex}
Susmit Jha, Ashish Tiwari, Sanjit~A Seshia, Tuhin Sahai, and Natarajan Shankar.
\newblock Telex: learning signal temporal logic from positive examples using
  tightness metric.
\newblock \emph{Formal Methods in System Design}, 54\penalty0 (3):\penalty0
  364--387, 2019.

\bibitem[Jiang et~al.(2020)Jiang, Bharadwaj, Wu, Shah, Topcu, and
  Stone]{jiang2020temporal}
Yuqian Jiang, Sudarshanan Bharadwaj, Bo~Wu, Rishi Shah, Ufuk Topcu, and Peter
  Stone.
\newblock Temporal-logic-based reward shaping for continuing learning tasks.
\newblock \emph{arXiv preprint arXiv:2007.01498}, 2020.

\bibitem[Jothimurugan et~al.(2019)Jothimurugan, Alur, and
  Bastani]{jothimurugan2019composable}
Kishor Jothimurugan, Rajeev Alur, and Osbert Bastani.
\newblock A composable specification language for reinforcement learning tasks.
\newblock \emph{Advances in Neural Information Processing Systems}, 32, 2019.

\bibitem[Jothimurugan et~al.(2021)Jothimurugan, Bansal, Bastani, and
  Alur]{jothimurugan2021compositional}
Kishor Jothimurugan, Suguman Bansal, Osbert Bastani, and Rajeev Alur.
\newblock Compositional reinforcement learning from logical specifications.
\newblock \emph{Advances in Neural Information Processing Systems},
  34:\penalty0 10026--10039, 2021.

\bibitem[Kingma \& Welling(2013)Kingma and Welling]{kingma2013auto}
Diederik~P Kingma and Max Welling.
\newblock Auto-encoding variational bayes.
\newblock \emph{arXiv preprint arXiv:1312.6114}, 2013.

\bibitem[Kloetzer \& Belta(2008)Kloetzer and Belta]{cbelta_ltl}
Marius Kloetzer and Calin Belta.
\newblock A fully automated framework for control of linear systems from
  temporal logic specifications.
\newblock \emph{IEEE Transactions on Automatic Control}, 53\penalty0
  (1):\penalty0 287--297, 2008.
\newblock \doi{10.1109/TAC.2007.914952}.

\bibitem[Kress-Gazit et~al.(2009)Kress-Gazit, Fainekos, and Pappas]{kg_ltl}
Hadas Kress-Gazit, Georgios~E. Fainekos, and George~J. Pappas.
\newblock Temporal-logic-based reactive mission and motion planning.
\newblock \emph{IEEE Transactions on Robotics}, 25\penalty0 (6):\penalty0
  1370--1381, 2009.
\newblock \doi{10.1109/TRO.2009.2030225}.

\bibitem[Leung et~al.(2022)Leung, Ar\'{e}chiga, and
  Pavone]{LeungArechigaEtAl2021}
K.~Leung, N.~Ar\'{e}chiga, and M.~Pavone.
\newblock Backpropagation through signal temporal logic specifications:
  Infusing logical structure into gradient-based methods.
\newblock \emph{{Int.\ Journal of Robotics Research}}, 2022.

\bibitem[Leung et~al.(2020)Leung, Ar{\'e}chiga, and Pavone]{leung2020back}
Karen Leung, Nikos Ar{\'e}chiga, and Marco Pavone.
\newblock Back-propagation through signal temporal logic specifications:
  Infusing logical structure into gradient-based methods.
\newblock In \emph{International Workshop on the Algorithmic Foundations of
  Robotics}, pp.\  432--449. Springer, 2020.

\bibitem[Levy et~al.(2017)Levy, Platt, and Saenko]{levy2017hierarchical}
Andrew Levy, Robert Platt, and Kate Saenko.
\newblock Hierarchical actor-critic.
\newblock \emph{arXiv preprint arXiv:1712.00948}, 12, 2017.

\bibitem[Li et~al.(2017)Li, Vasile, and Belta]{li2017reinforcement}
Xiao Li, Cristian-Ioan Vasile, and Calin Belta.
\newblock Reinforcement learning with temporal logic rewards.
\newblock In \emph{2017 IEEE/RSJ International Conference on Intelligent Robots
  and Systems (IROS)}, pp.\  3834--3839. IEEE, 2017.

\bibitem[Maler \& Nickovic(2004)Maler and Nickovic]{maler2004monitoring}
Oded Maler and Dejan Nickovic.
\newblock Monitoring temporal properties of continuous signals.
\newblock In \emph{Formal Techniques, Modelling and Analysis of Timed and
  Fault-Tolerant Systems}, pp.\  152--166. Springer, 2004.

\bibitem[Manhaeve et~al.(2018)Manhaeve, Dumancic, Kimmig, Demeester, and
  De~Raedt]{DeepProbLog}
Robin Manhaeve, Sebastijan Dumancic, Angelika Kimmig, Thomas Demeester, and Luc
  De~Raedt.
\newblock Deepproblog: Neural probabilistic logic programming.
\newblock In S.~Bengio, H.~Wallach, H.~Larochelle, K.~Grauman, N.~Cesa-Bianchi,
  and R.~Garnett (eds.), \emph{Advances in Neural Information Processing
  Systems}, volume~31. Curran Associates, Inc., 2018.
\newblock URL
  \url{https://proceedings.neurips.cc/paper/2018/file/dc5d637ed5e62c36ecb73b654b05ba2a-Paper.pdf}.

\bibitem[Nachum et~al.(2018)Nachum, Gu, Lee, and Levine]{nachum2018data}
Ofir Nachum, Shixiang~Shane Gu, Honglak Lee, and Sergey Levine.
\newblock Data-efficient hierarchical reinforcement learning.
\newblock \emph{Advances in neural information processing systems}, 31, 2018.

\bibitem[Panerati et~al.(2021)Panerati, Zheng, Zhou, Xu, Prorok, and
  Schoellig]{panerati2021learning}
Jacopo Panerati, Hehui Zheng, SiQi Zhou, James Xu, Amanda Prorok, and Angela~P.
  Schoellig.
\newblock Learning to fly---a gym environment with pybullet physics for
  reinforcement learning of multi-agent quadcopter control.
\newblock In \emph{2021 IEEE/RSJ International Conference on Intelligent Robots
  and Systems (IROS)}, 2021.

\bibitem[Park et~al.(2019)Park, Florence, Straub, Newcombe, and
  Lovegrove]{park2019deepsdf}
Jeong~Joon Park, Peter Florence, Julian Straub, Richard Newcombe, and Steven
  Lovegrove.
\newblock Deepsdf: Learning continuous signed distance functions for shape
  representation.
\newblock In \emph{Proceedings of the IEEE/CVF conference on computer vision
  and pattern recognition}, pp.\  165--174, 2019.

\bibitem[Pnueli(1977)]{first_ltl}
Amir Pnueli.
\newblock The temporal logic of programs.
\newblock In \emph{18th Annual Symposium on Foundations of Computer Science
  (sfcs 1977)}, pp.\  46--57, 1977.
\newblock \doi{10.1109/SFCS.1977.32}.

\bibitem[Qiu \& Zhu(2021)Qiu and Zhu]{qiu2021programmatic}
Wenjie Qiu and He~Zhu.
\newblock Programmatic reinforcement learning without oracles.
\newblock In \emph{International Conference on Learning Representations}, 2021.

\bibitem[Qureshi et~al.(2019)Qureshi, Simeonov, Bency, and
  Yip]{qureshi2019motion}
Ahmed~H Qureshi, Anthony Simeonov, Mayur~J Bency, and Michael~C Yip.
\newblock Motion planning networks.
\newblock In \emph{2019 International Conference on Robotics and Automation
  (ICRA)}, pp.\  2118--2124. IEEE, 2019.

\bibitem[Ratliff et~al.(2009)Ratliff, Zucker, Bagnell, and
  Srinivasa]{ratliff2009chomp}
Nathan Ratliff, Matt Zucker, J~Andrew Bagnell, and Siddhartha Srinivasa.
\newblock Chomp: Gradient optimization techniques for efficient motion
  planning.
\newblock In \emph{2009 IEEE International Conference on Robotics and
  Automation}, pp.\  489--494. IEEE, 2009.

\bibitem[Schaul et~al.(2015)Schaul, Horgan, Gregor, and
  Silver]{schaul2015universal}
Tom Schaul, Daniel Horgan, Karol Gregor, and David Silver.
\newblock Universal value function approximators.
\newblock In \emph{International conference on machine learning}, pp.\
  1312--1320. PMLR, 2015.

\bibitem[Schulman et~al.(2015)Schulman, Levine, Abbeel, Jordan, and
  Moritz]{schulman2015trust}
John Schulman, Sergey Levine, Pieter Abbeel, Michael Jordan, and Philipp
  Moritz.
\newblock Trust region policy optimization.
\newblock In \emph{International conference on machine learning}, pp.\
  1889--1897. PMLR, 2015.

\bibitem[Schulman et~al.(2017)Schulman, Wolski, Dhariwal, Radford, and
  Klimov]{schulman2017proximal}
John Schulman, Filip Wolski, Prafulla Dhariwal, Alec Radford, and Oleg Klimov.
\newblock Proximal policy optimization algorithms.
\newblock \emph{arXiv preprint arXiv:1707.06347}, 2017.

\bibitem[Shu et~al.(2018)Shu, Xiong, and Socher]{shu2018hierarchical}
Tianmin Shu, Caiming Xiong, and Richard Socher.
\newblock Hierarchical and interpretable skill acquisition in multi-task
  reinforcement learning.
\newblock In \emph{International Conference on Learning Representations}, 2018.
\newblock URL \url{https://openreview.net/forum?id=SJJQVZW0b}.

\bibitem[Stooke et~al.(2020)Stooke, Achiam, and Abbeel]{stooke2020responsive}
Adam Stooke, Joshua Achiam, and Pieter Abbeel.
\newblock Responsive safety in reinforcement learning by pid lagrangian
  methods.
\newblock In \emph{International Conference on Machine Learning}, pp.\
  9133--9143. PMLR, 2020.

\bibitem[Sutton et~al.(1999)Sutton, Precup, and Singh]{sutton1999between}
Richard~S Sutton, Doina Precup, and Satinder Singh.
\newblock Between mdps and semi-mdps: A framework for temporal abstraction in
  reinforcement learning.
\newblock \emph{Artificial intelligence}, 112\penalty0 (1-2):\penalty0
  181--211, 1999.

\bibitem[Wongpiromsarn et~al.(2012)Wongpiromsarn, Topcu, and
  Murray]{receding_ltl_planning}
Tichakorn Wongpiromsarn, Ufuk Topcu, and Richard~M. Murray.
\newblock Receding horizon temporal logic planning.
\newblock \emph{IEEE Transactions on Automatic Control}, 57\penalty0
  (11):\penalty0 2817--2830, 2012.
\newblock \doi{10.1109/TAC.2012.2195811}.

\bibitem[Xu et~al.(2020)Xu, Gavran, Ahmad, Majumdar, Neider, Topcu, and
  Wu]{xu2020joint}
Zhe Xu, Ivan Gavran, Yousef Ahmad, Rupak Majumdar, Daniel Neider, Ufuk Topcu,
  and Bo~Wu.
\newblock Joint inference of reward machines and policies for reinforcement
  learning.
\newblock \emph{Proceedings of the International Conference on Automated
  Planning and Scheduling}, 30\penalty0 (1):\penalty0 590--598, Jun. 2020.
\newblock URL \url{https://ojs.aaai.org/index.php/ICAPS/article/view/6756}.

\bibitem[Yang et~al.(2022)Yang, Littman, and Carbin]{yang2021reinforcement}
Cambridge Yang, Michael~L. Littman, and Michael Carbin.
\newblock On the (in)tractability of reinforcement learning for {LTL}
  objectives.
\newblock In Luc~De Raedt (ed.), \emph{Proceedings of the Thirty-First
  International Joint Conference on Artificial Intelligence, {IJCAI} 2022,
  Vienna, Austria, 23-29 July 2022}, pp.\  3650--3658. ijcai.org, 2022.
\newblock \doi{10.24963/ijcai.2022/507}.
\newblock URL \url{https://doi.org/10.24963/ijcai.2022/507}.

\bibitem[Yang et~al.(2021)Yang, Inala, Bastani, Pu, Solar-Lezama, and
  Rinard]{yang2021program}
Yichen Yang, Jeevana~Priya Inala, Osbert Bastani, Yewen Pu, Armando
  Solar-Lezama, and Martin Rinard.
\newblock Program synthesis guided reinforcement learning for partially
  observed environments.
\newblock In M.~Ranzato, A.~Beygelzimer, Y.~Dauphin, P.S. Liang, and J.~Wortman
  Vaughan (eds.), \emph{Advances in Neural Information Processing Systems},
  volume~34, pp.\  29669--29683. Curran Associates, Inc., 2021.
\newblock URL
  \url{https://proceedings.neurips.cc/paper/2021/file/f7e2b2b75b04175610e5a00c1e221ebb-Paper.pdf}.

\bibitem[Yao et~al.(2014)Yao, Szepesvari, Sutton, Modayil, and
  Bhatnagar]{NIPS2014_996a7fa0}
Hengshuai Yao, Csaba Szepesvari, Richard~S Sutton, Joseph Modayil, and Shalabh
  Bhatnagar.
\newblock Universal option models.
\newblock In Z.~Ghahramani, M.~Welling, C.~Cortes, N.~Lawrence, and K.Q.
  Weinberger (eds.), \emph{Advances in Neural Information Processing Systems},
  volume~27. Curran Associates, Inc., 2014.
\newblock URL
  \url{https://proceedings.neurips.cc/paper/2014/file/996a7fa078cc36c46d02f9af3bef918b-Paper.pdf}.

\bibitem[Zhang \& Kan(2022)Zhang and Kan]{zhang2022temporal}
Hao Zhang and Zhen Kan.
\newblock Temporal logic guided meta q-learning of multiple tasks.
\newblock \emph{IEEE Robotics and Automation Letters}, 7\penalty0 (3):\penalty0
  8194--8201, 2022.
\newblock \doi{10.1109/LRA.2022.3185384}.

\bibitem[Zhang et~al.(2020)Zhang, Guo, Tan, Hu, and Chen]{zhang2020generating}
Tianren Zhang, Shangqi Guo, Tian Tan, Xiaolin Hu, and Feng Chen.
\newblock Generating adjacency-constrained subgoals in hierarchical
  reinforcement learning.
\newblock In H.~Larochelle, M.~Ranzato, R.~Hadsell, M.F. Balcan, and H.~Lin
  (eds.), \emph{Advances in Neural Information Processing Systems}, volume~33,
  pp.\  21579--21590. Curran Associates, Inc., 2020.
\newblock URL
  \url{https://proceedings.neurips.cc/paper/2020/file/f5f3b8d720f34ebebceb7765e447268b-Paper.pdf}.

\bibitem[Ziebart et~al.(2008)Ziebart, Maas, Bagnell, Dey,
  et~al.]{ziebart2008maximum}
Brian~D Ziebart, Andrew~L Maas, J~Andrew Bagnell, Anind~K Dey, et~al.
\newblock Maximum entropy inverse reinforcement learning.
\newblock In \emph{Aaai}, volume~8, pp.\  1433--1438. Chicago, IL, USA, 2008.

\end{thebibliography}


\begin{thebibliography}{10}
\providecommand{\url}[1]{#1}
\csname url@rmstyle\endcsname
\providecommand{\newblock}{\relax}
\providecommand{\bibinfo}[2]{#2}
\providecommand\BIBentrySTDinterwordspacing{\spaceskip=0pt\relax}
\providecommand\BIBentryALTinterwordstretchfactor{4}
\providecommand\BIBentryALTinterwordspacing{\spaceskip=\fontdimen2\font plus
\BIBentryALTinterwordstretchfactor\fontdimen3\font minus
  \fontdimen4\font\relax}
\providecommand\BIBforeignlanguage[2]{{%
\expandafter\ifx\csname l@#1\endcsname\relax
\typeout{** WARNING: IEEEtran.bst: No hyphenation pattern has been}%
\typeout{** loaded for the language `#1'. Using the pattern for}%
\typeout{** the default language instead.}%
\else
\language=\csname l@#1\endcsname
\fi
#2}}

\bibitem{fu2014probably}
J.~Fu and U.~Topcu, ``Probably approximately correct mdp learning and control
  with temporal logic constraints,'' \emph{arXiv preprint arXiv:1404.7073},
  2014.

\bibitem{li2017reinforcement}
X.~Li, C.-I. Vasile, and C.~Belta, ``Reinforcement learning with temporal logic
  rewards,'' in \emph{2017 IEEE/RSJ International Conference on Intelligent
  Robots and Systems (IROS)}.\hskip 1em plus 0.5em minus 0.4em\relax IEEE,
  2017, pp. 3834--3839.

\bibitem{hasanbeig2019reinforcement}
M.~Hasanbeig, Y.~Kantaros, A.~Abate, D.~Kroening, G.~J. Pappas, and I.~Lee,
  ``Reinforcement learning for temporal logic control synthesis with
  probabilistic satisfaction guarantees,'' in \emph{2019 IEEE 58th Conference
  on Decision and Control (CDC)}.\hskip 1em plus 0.5em minus 0.4em\relax IEEE,
  2019, pp. 5338--5343.

\bibitem{jothimurugan2019composable}
K.~Jothimurugan, R.~Alur, and O.~Bastani, ``A composable specification language
  for reinforcement learning tasks,'' \emph{Advances in Neural Information
  Processing Systems}, vol.~32, 2019.

\bibitem{icarte2022reward}
R.~T. Icarte, T.~Q. Klassen, R.~Valenzano, and S.~A. McIlraith, ``Reward
  machines: Exploiting reward function structure in reinforcement learning,''
  \emph{Journal of Artificial Intelligence Research}, vol.~73, pp. 173--208,
  2022.

\bibitem{lcrl_tool}
M.~Hasanbeig, D.~Kroening, and A.~Abate, ``{LCRL}: Certified policy synthesis
  via logically-constrained reinforcement learning,'' in \emph{International
  Conference on Quantitative Evaluation of SysTems}.\hskip 1em plus 0.5em minus
  0.4em\relax Springer, 2022.

\bibitem{jiang2020temporal}
Y.~Jiang, S.~Bharadwaj, B.~Wu, R.~Shah, U.~Topcu, and P.~Stone,
  ``Temporal-logic-based reward shaping for continuing learning tasks,''
  \emph{arXiv preprint arXiv:2007.01498}, 2020.

\bibitem{bozkurt2020control}
A.~K. Bozkurt, Y.~Wang, M.~M. Zavlanos, and M.~Pajic, ``Control synthesis from
  linear temporal logic specifications using model-free reinforcement
  learning,'' in \emph{2020 IEEE International Conference on Robotics and
  Automation (ICRA)}.\hskip 1em plus 0.5em minus 0.4em\relax IEEE, 2020, pp.
  10\,349--10\,355.

\bibitem{xu2020joint}
Z.~Xu, I.~Gavran, Y.~Ahmad, R.~Majumdar, D.~Neider, U.~Topcu, and B.~Wu,
  ``Joint inference of reward machines and policies for reinforcement
  learning,'' \emph{Proceedings of the International Conference on Automated
  Planning and Scheduling}, vol.~30, no.~1, pp. 590--598, Jun. 2020.

\bibitem{zhang2022temporal}
H.~Zhang and Z.~Kan, ``Temporal logic guided meta q-learning of multiple
  tasks,'' \emph{IEEE Robotics and Automation Letters}, vol.~7, no.~3, pp.
  8194--8201, 2022.

\bibitem{dulac2019challenges}
G.~Dulac-Arnold, D.~Mankowitz, and T.~Hester, ``Challenges of real-world
  reinforcement learning,'' \emph{arXiv preprint arXiv:1904.12901}, 2019.

\bibitem{kurtz2022mixed}
V.~Kurtz and H.~Lin, ``Mixed-integer programming for signal temporal logic with
  fewer binary variables,'' \emph{IEEE Control Systems Letters}, 2022.

\bibitem{first_ltl}
A.~Pnueli, ``The temporal logic of programs,'' in \emph{18th Annual Symposium
  on Foundations of Computer Science (sfcs 1977)}, 1977, pp. 46--57.

\bibitem{maler2004monitoring}
O.~Maler and D.~Nickovic, ``Monitoring temporal properties of continuous
  signals,'' in \emph{Formal Techniques, Modelling and Analysis of Timed and
  Fault-Tolerant Systems}.\hskip 1em plus 0.5em minus 0.4em\relax Springer,
  2004, pp. 152--166.

\bibitem{fainekos2005temporal}
G.~E. Fainekos, H.~Kress-Gazit, and G.~J. Pappas, ``Temporal logic motion
  planning for mobile robots,'' in \emph{Proceedings of the 2005 IEEE
  International Conference on Robotics and Automation}.\hskip 1em plus 0.5em
  minus 0.4em\relax IEEE, 2005, pp. 2020--2025.

\bibitem{cbelta_ltl}
M.~Kloetzer and C.~Belta, ``A fully automated framework for control of linear
  systems from temporal logic specifications,'' \emph{IEEE Transactions on
  Automatic Control}, vol.~53, no.~1, pp. 287--297, 2008.

\bibitem{kg_ltl}
H.~Kress-Gazit, G.~E. Fainekos, and G.~J. Pappas, ``Temporal-logic-based
  reactive mission and motion planning,'' \emph{IEEE Transactions on Robotics},
  vol.~25, no.~6, pp. 1370--1381, 2009.

\bibitem{receding_ltl_planning}
T.~Wongpiromsarn, U.~Topcu, and R.~M. Murray, ``Receding horizon temporal logic
  planning,'' \emph{IEEE Transactions on Automatic Control}, vol.~57, no.~11,
  pp. 2817--2830, 2012.

\bibitem{yang2021reinforcement}
C.~Yang, M.~L. Littman, and M.~Carbin, ``On the (in)tractability of
  reinforcement learning for {LTL} objectives,'' in \emph{Proceedings of the
  Thirty-First International Joint Conference on Artificial Intelligence,
  {IJCAI} 2022}, 2022, pp. 3650--3658.

\bibitem{schulman2015trust}
J.~Schulman, S.~Levine, P.~Abbeel, M.~Jordan, and P.~Moritz, ``Trust region
  policy optimization,'' in \emph{International conference on machine
  learning}.\hskip 1em plus 0.5em minus 0.4em\relax PMLR, 2015, pp. 1889--1897.

\bibitem{achiam2017constrained}
J.~Achiam, D.~Held, A.~Tamar, and P.~Abbeel, ``Constrained policy
  optimization,'' in \emph{International conference on machine learning}.\hskip
  1em plus 0.5em minus 0.4em\relax PMLR, 2017, pp. 22--31.

\bibitem{schulman2017proximal}
J.~Schulman, F.~Wolski, P.~Dhariwal, A.~Radford, and O.~Klimov, ``Proximal
  policy optimization algorithms,'' \emph{arXiv preprint arXiv:1707.06347},
  2017.

\bibitem{LeungArechigaEtAl2021}
K.~Leung, N.~Ar\'{e}chiga, and M.~Pavone, ``Backpropagation through signal
  temporal logic specifications: Infusing logical structure into gradient-based
  methods,'' \emph{{Int.\ Journal of Robotics Research}}, 2022.

\bibitem{leung2020back}
K.~Leung, N.~Ar{\'e}chiga, and M.~Pavone, ``Back-propagation through signal
  temporal logic specifications: Infusing logical structure into gradient-based
  methods,'' in \emph{International Workshop on the Algorithmic Foundations of
  Robotics}.\hskip 1em plus 0.5em minus 0.4em\relax Springer, 2020, pp.
  432--449.

\bibitem{ratliff2009chomp}
N.~Ratliff, M.~Zucker, J.~A. Bagnell, and S.~Srinivasa, ``Chomp: Gradient
  optimization techniques for efficient motion planning,'' in \emph{2009 IEEE
  International Conference on Robotics and Automation}.\hskip 1em plus 0.5em
  minus 0.4em\relax IEEE, 2009, pp. 489--494.

\bibitem{campana2016gradient}
M.~Campana, F.~Lamiraux, and J.-P. Laumond, ``A gradient-based path
  optimization method for motion planning,'' \emph{Advanced Robotics}, vol.~30,
  no. 17-18, pp. 1126--1144, 2016.

\bibitem{dawson2022robust}
C.~Dawson and C.~Fan, ``Robust counterexample-guided optimization for planning
  from differentiable temporal logic,'' \emph{arXiv preprint arXiv:2203.02038},
  2022.

\bibitem{qureshi2019motion}
A.~H. Qureshi, A.~Simeonov, M.~J. Bency, and M.~C. Yip, ``Motion planning
  networks,'' in \emph{2019 International Conference on Robotics and Automation
  (ICRA)}.\hskip 1em plus 0.5em minus 0.4em\relax IEEE, 2019, pp. 2118--2124.

\bibitem{williams1992simple}
R.~J. Williams, ``Simple statistical gradient-following algorithms for
  connectionist reinforcement learning,'' \emph{Machine learning}, vol.~8, pp.
  229--256, 1992.

\bibitem{wu2018variance}
C.~Wu, A.~Rajeswaran, Y.~Duan, V.~Kumar, A.~M. Bayen, S.~Kakade, I.~Mordatch,
  and P.~Abbeel, ``Variance reduction for policy gradient with action-dependent
  factorized baselines,'' \emph{arXiv preprint arXiv:1803.07246}, 2018.

\bibitem{smoothStl}
Y.~Gilpin, V.~Kurtz, and H.~Lin, ``A smooth robustness measure of signal
  temporal logic for symbolic control,'' \emph{IEEE Control Systems Letters},
  vol.~5, no.~1, pp. 241--246, 2021.

\bibitem{stooke2020responsive}
A.~Stooke, J.~Achiam, and P.~Abbeel, ``Responsive safety in reinforcement
  learning by pid lagrangian methods,'' in \emph{International Conference on
  Machine Learning}.\hskip 1em plus 0.5em minus 0.4em\relax PMLR, 2020, pp.
  9133--9143.

\bibitem{zhang2023faster}
C.~Zhang, D.~Han, Y.~Qiao, J.~U. Kim, S.-H. Bae, S.~Lee, and C.~S. Hong,
  ``Faster segment anything: Towards lightweight sam for mobile applications,''
  \emph{arXiv preprint arXiv:2306.14289}, 2023.

\bibitem{mehta2021mobilevit}
S.~Mehta and M.~Rastegari, ``Mobilevit: light-weight, general-purpose, and
  mobile-friendly vision transformer,'' \emph{arXiv preprint arXiv:2110.02178},
  2021.

\bibitem{vaswani2017attention}
A.~Vaswani, N.~Shazeer, N.~Parmar, J.~Uszkoreit, L.~Jones, A.~N. Gomez,
  {\L}.~Kaiser, and I.~Polosukhin, ``Attention is all you need,''
  \emph{Advances in neural information processing systems}, vol.~30, 2017.

\bibitem{malladi1995shape}
R.~Malladi, J.~A. Sethian, and B.~C. Vemuri, ``Shape modeling with front
  propagation: A level set approach,'' \emph{IEEE transactions on pattern
  analysis and machine intelligence}, vol.~17, no.~2, pp. 158--175, 1995.

\bibitem{kdtree}
\BIBentryALTinterwordspacing
J.~L. Bentley, ``Multidimensional binary search trees used for associative
  searching,'' \emph{Commun. ACM}, vol.~18, no.~9, p. 509–517, sep 1975.
  [Online]. Available: \url{https://doi.org/10.1145/361002.361007}
\BIBentrySTDinterwordspacing

\bibitem{schaul2015universal}
T.~Schaul, D.~Horgan, K.~Gregor, and D.~Silver, ``Universal value function
  approximators,'' in \emph{International conference on machine
  learning}.\hskip 1em plus 0.5em minus 0.4em\relax PMLR, 2015, pp. 1312--1320.

\end{thebibliography}
